\def\year{2021}\relax
\documentclass[letterpaper]{article} 
\usepackage{aaai21}  
\usepackage{times}  
\usepackage{helvet} 
\usepackage{courier}  
\usepackage[hyphens]{url}  
\usepackage{graphicx} 
\usepackage{natbib}  
\usepackage{caption} 
\usepackage{booktabs}       
\usepackage{amsfonts}       
\usepackage{subfigure}
\usepackage{multirow}
\usepackage{tikz}
\usepackage{pgfplots}
\pgfplotsset{compat=1.14}
\usepackage{amsmath,amssymb} 

\DeclareMathOperator*{\argmin}{argmin}

\title{Stochastic Precision Ensemble: Self-Knowledge Distillation for Quantized Deep Neural Networks}
\author {
    Yoonho Boo,\textsuperscript{\rm 1}
    Sungho Shin,\textsuperscript{\rm 1}
    Jungwook Choi, \textsuperscript{\rm 2}
    and Wonyong Sung \textsuperscript{\rm 1, \rm 3} \\
}
\affiliations {
    \textsuperscript{\rm 1} Seoul National University, Seoul, Korea \\
    \textsuperscript{\rm 2} Hanyang University, Seoul, Korea \\
    \textsuperscript{\rm 3} Gwangju Institue of Science and Technology (GIST), Gwangju, Korea \\
    dnsgh337@snu.ac.kr, sungho.develop@gmail.com, choij@hanyang.ac.kr, wysung@snu.ac.kr
}

\begin{document}

\maketitle

\begin{abstract}
The quantization of deep neural networks (QDNNs) has been actively studied for deployment in edge devices. Recent studies employ the knowledge distillation (KD) method to improve the performance of quantized networks. In this study, we propose stochastic precision ensemble training for QDNNs (SPEQ). SPEQ is a knowledge distillation training scheme; however, the teacher is formed by sharing the model parameters of the student network. We obtain the soft labels of the teacher by changing the bit precision of the activation stochastically at each layer of the forward-pass computation. The student model is trained with these soft labels to reduce the activation quantization noise. The cosine similarity loss is employed, instead of the KL-divergence, for KD training. As the teacher model changes continuously by random bit-precision assignment, it exploits the effect of stochastic ensemble KD. SPEQ outperforms the existing quantization training methods in various tasks, such as image classification, question-answering, and transfer learning without the need for cumbersome teacher networks.
\end{abstract}

\section{Introduction}
\label{ch4:sec:intro}

Deep neural networks (DNNs) have achieved remarkable accuracy for tasks in a wide range of applications, including image processing \cite{he2016deep}, machine translation \cite{gehring2017convolutional}, and speech recognition \cite{zhang2017towards}. These state-of-the-art neural networks use very deep models, consuming hundreds of ExaOps of computation during training and GBytes of storage for model and data. This complexity poses a tremendous challenge for widespread deployment, especially in resource-constrained edge environments, leading to a plethora of explorations in model compression that minimize memory footprint and computational complexity while attempting to preserve the performance of the model. Among them, research on quantized DNNs (QDNNs) focuses on quantizing key data structures, namely weights and activations, into low-precision. Hence, we can save memory access overhead and simplify the arithmetic unit to perform reduced-precision computation. There have been extensive studies on QDNNs~\cite{fengfu2016ternary,courbariaux2015binaryconnect,choi2018pact,hou2018loss}, but most of them suffer from accuracy loss due to quantization.


To enhance the performance of low-capacity models, knowledge distillation~\cite{hinton2015distilling,bucilua2006model} (KD) has been widely adopted. KD employs a more accurate model as a teacher network to guide the training of a student model. For the same input, the teacher network provides its prediction as a soft label, which can be further considered in the loss function to guide the training of the student network. In the case of QDNNs, the quantized student network can compensate for its accuracy loss via supervision of the teacher model~\cite{mishra2018apprentice,polino2018model,shin2019empirical,kim2019qkd}. However, the need for large and high-performance teacher models introduces significant overhead when applying KD. In particular, KD has not been successfully employed in the emerging study of on-device training for model adaptation and transfer learning, since the memory-intensive teacher models may not be available once the quantized models are deployed.



In this work, we propose a new practical approach to KD for QDNNs, called stochastic precision ensemble training for QDNNs (SPEQ). SPEQ is motivated by an inspiring observation about activation quantization. \tablename~\ref{ch4:table:lowtohighex} shows that the accuracy of the WFA2 (float weight and 2-bit activation) model improves as the activation precision increases. However, the W2AF (2-bit weight and float activation) model shows the opposite characteristic. The accuracy drops as the weight precision increases for inference. This simple experiment reveals interesting insights: the activation quantization mostly adds noise to the decision boundary~\cite{boo2020quantized}. Therefore, inference with various activation precision results in selective removal of such noise, leading to diverse guidance that can be exploited for self knowledge distillation.




\begin{table*}[t]
\setlength\tabcolsep{5pt}
\caption{CIFAR100 test accuracy (\%) in higher precision on the quantized model. ResNet20 is trained with 2-bit weight / float activation and float weight / 2-bit activation. (Details in Appendix A.)}
\label{ch4:table:lowtohighex}
\begin{center}
\begin{tabular}{c|cccc}
\toprule
Trained precision &
\multicolumn{4}{c}{Test accuracy (\%) / Inference precision} \\ \midrule
2-bit W, float A (W2AF)                   & 65.74 / W2AF    & 58.01 / W4AF    & 55.85 / W8AF   & 54.70 / WFAF       \\
Float W, 2-bit A (WFA2)                   & 66.93 / WFA2    & 68.48 / WFA4    & 68.77 / WFA8   & 68.71 / WFAF   \\
\bottomrule
\end{tabular}
\end{center}
\end{table*}

In SPEQ, we form a teacher network that shares the quantized weights with the student but employs different bit precision for activation. The clipping levels of activation are also shared. In fact, the activation precision for the teacher is randomly selected between the low and high precision, such as 2 and 8-bit. Since the teacher stochastically applies the target low-bit activation quantization for its soft label computation, it can experience the impact of quantization for the guidance. Furthermore, we reveal that the cosine similarity loss is essential for distilling the knowledge of the teacher of stochastic quantization to the low-precision student. 

Although this form of guidance resembles KD, there is a significant difference in that the same model is shared and any other auxiliary models, such as large teacher networks, are unnecessary. The forward-pass computation of the teacher and student in SPEQ can be performed economically as the same weight parameters can be loaded only once. Therefore, the SPEQ can improve the performance much without the overhead of teacher-model search or hyper-parameter tuning needed for conventional KD. Furthermore, since the stochastic precision ensemble provides distinctive knowledge, SPEQ can be combined with the conventional KD method to further improve the performance of the target QDNNs.


We demonstrate the superior performance and efficiency of our SPEQ on various applications, including CIFAR10/CIFAR100/ImageNet image classification and also transfer learning scenarios such as BERT-based question-answering and flower classification.

The contributions of our work are summarized as follows:
\begin{itemize}
\itemsep0em
\item We propose a new practical KD method called SPEQ that can enhance the accuracy of QDNNs QDNNs employing low-precision bit-widths for weights and activation signals. This method can yield better results compared to conventional KD-based QDNN optimization that utilizes large teacher models.
\item We suggest cosine similarity as an essential loss function to effectively distill the knowledge of activation quantization in SPEQ training. 
\item We demonstrate that the proposed method outperforms the existing KD methods for training QDNNs with lower training overhead. We confirm this on various models and tasks including image classifications, question answering, and transfer learning. 
\item We show that the proposed method can be combined to the conventional KD method with a large teacher to further improve the performance of the target model.
\end{itemize}

\section{Related Works}
\label{ch4:sec:related}

\subsection{Quantization of Deep Neural Networks}
\label{ch4:subsec:quantize_char}
QDNNs have been studied for a long time. Early works suggested stochastic gradient descent (SGD)-based training for QDNNs to restore the performance reduced by the quantization error~\cite{courbariaux2015binaryconnect,hwang2014fixed,zhu2016trained}. The quantized SGD training maintains both full-precision and quantized weights. Full-precision weights are exploited to accumulate the gradients, and the quantized weights are used for computing forward and backward propagation. Several techniques have been combined with the quantized SGD algorithm, which include data distribution \cite{zhou2017balanced}, stochastic rounding \cite{gupta2015deep}, weight cluster \cite{park2017weighted}, trainable quantization \cite{zhang2018lq}, fittable quantization scale \cite{cai2017deep}, pow2-ternaization~\cite{ott2016recurrent}, stochastic weight averaging~\cite{shin2020sqwa}, increasing the size of the neural network~\cite{kapur2017low}, and quantization interval learning~\cite{jung2019learning}. Recent study suggested that quantization errors for weight and activation are different~\cite{boo2020quantized}. Activation quantized models are known to be vulnerable to the adversarial noise~\cite{lin2019defensive}. Architectural modifications of increasing the width or moving the location of activation and batch normalization have also been studied~\cite{zagoruyko2016wide,he2016identity}. In particular, increasing the number of parameters in CNNs reduces the quantization sensitivity \cite{mishra2017wrpn}. However, considering the purpose of model compression, the number of parameters needs to be constrained.

\begin{figure*}[t]
\begin{center}
\centerline{\includegraphics[width=0.85\linewidth]{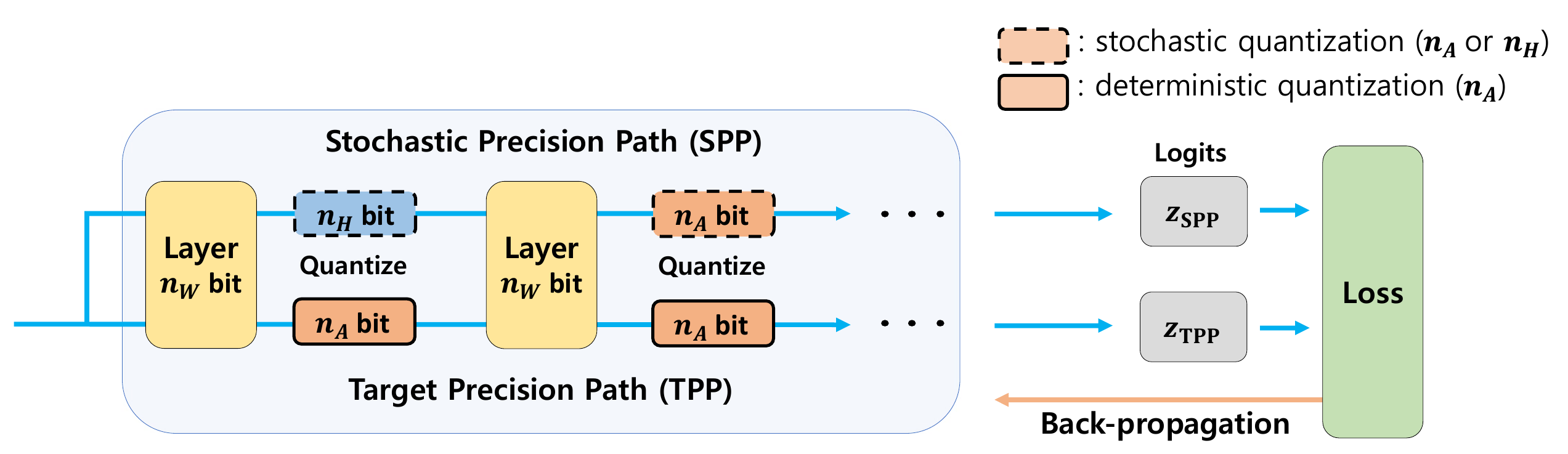}}
\caption{Structure of the proposed SPEQ training scheme for QDNNs. The QDNNs are trained for the target precision $n_{A}$ through the `target precision path'. The `stochastic precision path' produces the teacher logits, $\mathbf{z}_{\textrm{SPP}}$ using the same model but with randomly assigned quantization precision for activation at every iteration. Note that the weights in the model are quantized to $n_{W}$ bits.}
\label{ch4:fig:model}
\end{center}
\end{figure*}


\subsection{Knowledge Distillation for Quantization}
\label{ch4:secsec_related_KD}

KD is a method to improve the accuracy of a target model (called a student) by transferring better representation power (i.e., "knowledge") of a larger or more complex model (called a teacher)~\cite{hinton2015distilling,bucilua2006model}. Recently, several papers have adopted KD to restore the accuracy loss due to the quantization error of reduced-precision inference~\cite{zhuang2018towards,polino2018model,mishra2018apprentice,shin2019empirical,kim2019qkd}. Apprentice~\cite{mishra2018apprentice} proposed several approaches to apply KD for enhancing the accuracy of the quantized models. The importance of the hyperparameters of KD was studied in \cite{shin2019empirical}. More recently, quantization aware KD~\cite{kim2019qkd} (QKD) has been suggested, wherein the three training phases are coordinated as self-studying, co-studying, and tutoring. They train the full-precision larger teacher model using the soft labels of quantized student network to make the teacher understand the quantization errors of the student model. SP-Net~\cite{guerra2020switchable} also adopted self distillation but it focused on training models to robustly operate at various precisions. They employed only the full-precision pass as the teacher to improve accuracy for the other precision settings. Several studies have adopted deterministic self-distillation~\cite{zhang2019your,li2019improved,phuong2019distillation,yu2019universally,zhang2019scan}.

The difference between the proposed method and the previous works is that the teacher is formed by sharing the student model and assigning stochastic bit precision to activation. There are two main advantages of this method: the teacher information contains the quantization noise induced in the target QDNN by model sharing (better performance) and pretrained teacher models or auxiliary training parameters are unnecessary (lower training cost).



\section{Stochastic Precision Ensemble Training for QDNNs}
\label{ch4:sec:SPEQ}

\subsection{Stochastic Precision Self-Distillation with Model Sharing}

Changing the activation quantization precision in the same model affects the amount of noise injected into the model, as shown in~\tablename~\ref{ch4:table:lowtohighex}. That is, the outputs obtained through high-precision activation have information when the model is operated without noise. 



The training procedure of the SPEQ is illustrated in~\figurename~\ref{ch4:fig:model}. Two outputs are computed through different paths using the same parameters. Note that the initial quantized weights and clipping levels for activation are determined using conventional QDNN optimization methods~\cite{jung2019learning,choi2018pact}. The details of the employed quantization method are shown in Appendix B. The first output logits, $\mathbf{z}_{\textrm{TPP}}$, are obtained through the target precision path by quantizing the activation outputs to $n_{A}$ bits. The goal of the SPEQ is to increase the performance of the QDNN with this target precision path. The second output logits, $\mathbf{z}_{\textrm{SPP}}$, are computed by quantizing the activation outputs using the stochastic bit precision, $n_{\textrm{SPP}}$ which is defined as follows:
\begin{align}
    n_{\textrm{SPP}}^l =
    \left\{
        \begin{array}{ccc}
                n_{A} & \textrm{with probability } & u \\
                n_{H} & \textrm{with probability } & 1-u, \\
        \end{array} 
    \right. 
\end{align}
where $l$ denotes the layer index, $n_{A}$ is the target precision, $n_{H}$ is a precision higher than the target precision, and $u$ is a quantization probability for the stochastic quantization path. We set the high precision, $n_{H}$, to 8 bits. The impact of $u$ is discussed in the next section. For readability, we denote the set of $n_{\textrm{SPP}}^l$ as $\mathbf{n}_{\textrm{SPP}}$, that is, $\mathbf{n}_{\textrm{SPP}} = \{n_{\textrm{SPP}}^{1}, ... n_{\textrm{SPP}}^{L}\}$. 

The output probability, $\mathbf{p}(\mathbf(z),\mathcal{T})$, is computed using the softmax operation with temperature, $\mathcal{T}$. The temperature softens the distribution of the softmax outputs by dividing the output logits~\cite{hinton2015distilling}. Note that the soft labels, $\mathbf{p}(\mathbf{z}_{\textrm{SPP}}, \mathcal{T})$, are produced while sharing the parameters. Thus, they contain information of the quantization noise of the target model. We train the QDNN using these soft labels to reduce the activation quantization noise. The loss for the SPEQ training is the sum of the cross-entropy loss, $\mathcal{L}_{CE}$, and the cosine similarity loss, $\mathcal{L}_{CS}$, as follows. 
\begin{align}
    \mathcal{L}_{SPEQ} &= \mathcal{L}_{CE} (y, \mathbf{p}(\mathbf{z}_{\textrm{TPP}}, 1) ) \nonumber \\ 
    &+ \mathcal{L}_{CS} (\mathbf{p}(\mathbf{z}_{\textrm{SPP}}, \mathcal{T}), \mathbf{p}(\mathbf{z}_{\textrm{TPP}}, \mathcal{T}) ) \times \mathcal{T}^{2}. \label{ch4:eq:lossterm}
\end{align}
The effects of the cosine similarity loss function are discussed in Section~\ref{ch4:subsec:coslearing}. Note that the $\mathbf{z}_{\textrm{SPP}}$ is only used to produce the soft label for the $\mathcal{L}_{CS}$, thus, the back-propagation error only flows through the target precision path, as shown in~\figurename~\ref{ch4:fig:model}. Therefore, the only computational overhead for a training step is the computation of $\mathbf{z}_{\textrm{SPP}}$ by forward propagation. The SPEQ is based on the KD training but has the advantage that no other auxiliary model is required.

The proposed method can also adopt a larger teacher model to further improve the performance. In this case, the outputs through the SPP can be seen as the outputs of the teacher-assistant~\cite{mirzadeh2019improved}. The training loss with a larger teacher is computed as follows:
\begin{align}
\mathcal{L}_{\textrm{T}} &= \mathcal{L}_{KL} (\mathbf{p}(\mathbf{z}_{\textrm{T}}, \mathcal{T}), \mathbf{p}(\mathbf{z}_{\textrm{TPP}}, \mathcal{T}) ) \times \mathcal{T}^{2} \\
\mathcal{L}_{SPEQ+KD} &= \lambda \mathcal{L}_{SPEQ} + (1-\lambda) \mathcal{L}_{T}.
\end{align}
$\mathbf{z}_{\textrm{T}}$ is the logit from a large teacher model and thus $\mathcal{L}_{T}$ represents the distllation loss from the larger teacher to the shared model with the target precision.




\begin{figure}[t]
\centering
\subfigure[]{
\begin{tikzpicture}
    \begin{axis}[
	width=0.9\linewidth,
	height = 0.5\linewidth,
	xmin=0,
	ymin=0.4,
	xmax=175,
	ymax=1,
	label style={font=\footnotesize},
	legend style={nodes={scale=0.62, transform shape},at={(0.6,1.0)},anchor=north west},
	tick label style={font=\footnotesize}, 
	minor x tick num=4, 
	minor y tick num=4, 
	xtick pos=both, 
	xtick align=inside, 
	major tick style={line width=0.010cm, black},
	 major tick length=0.10cm,
        xlabel=Epoch,
        ylabel=8-bit Ratio]
	\legend{Conv1, Conv2, Conv3, Conv4, Conv5};
    \addplot[color=orange, mark=*, mark size=1.0pt, solid, mark repeat=15,mark options=solid] file{chapter3_sources/p_8bit_layer1.txt}; 
	\addplot[color=red!70!green, mark=Mercedes star flipped, mark size=2.5pt, solid, mark repeat=10,mark options=solid] file{chapter3_sources/p_8bit_layer2.txt}; 
	\addplot[color=red, mark=otimes, mark size=1.6pt, solid, mark repeat=15,mark options=solid] file{chapter3_sources/p_8bit_layer3.txt};
	\addplot[color=yellow!60!black, mark=square, mark size=1.6pt, solid, mark repeat=15,mark options=solid] file{chapter3_sources/p_8bit_layer4.txt}; 
	\addplot[color=blue!60!yellow, mark=diamond, mark size=1.6pt, solid, mark repeat=15,mark options=solid] file{chapter3_sources/p_8bit_layer5.txt};
    \end{axis}
   \end{tikzpicture}}
   \vskip 0.1in
   \subfigure[]{\begin{tikzpicture}
    \begin{axis}[
    ybar=0pt,
    width=0.9\linewidth,
	height = 0.5\linewidth,
	bar width=2.5pt,
    xmin=0,
	ymin=0,
	xmax=9,
	ymax=0.7,
    label style={font=\footnotesize},
    legend style={nodes={scale=0.62, transform shape},at={(0.8,0.99)},anchor=north west},
    xlabel={Class},
    ylabel={Softmax},
    tick label style={font=\footnotesize}, 
	minor x tick num=1, 
	enlarge x limits={abs=2*\pgfplotbarwidth},
	xtick pos=both, 
	xtick align=inside, 
	xtick={0, 1, 2, 3, 4, 5, 6, 7, 8, 9},
	major tick style={line width=0.010cm, black},
	major tick length=0.10cm,
	nodes near coords align={vertical},
    symbolic x coords={0,1,2,3,4,5,6,7,8,9}]
	\legend{ $n_D 1$, $n_D 2$, $n_D 3$, $n_D 4$, $n_D 5$};
    \addplot[color=blue!60!yellow,fill=blue!60!yellow] file{chapter3_sources/model0.txt}; 
	\addplot[color=red!70!green,fill=red!70!green] file{chapter3_sources/model1.txt}; 
 	\addplot[color=red,fill=red] file{chapter3_sources/model2.txt};
 	\addplot[color=yellow!60!black,fill=yellow!60!black] file{chapter3_sources/model3.txt}; 
 	\addplot[color=orange,fill=orange] file{chapter3_sources/model4.txt};
    \end{axis}
   \end{tikzpicture}}
\caption{(a): The ratio of selected 8-bit precision when trained with the greedy strategy. (b) Softmax distributions generated by the stochastic precision path with the same images.} \label{ch4:fig:ratio_change_curve}
\end{figure}
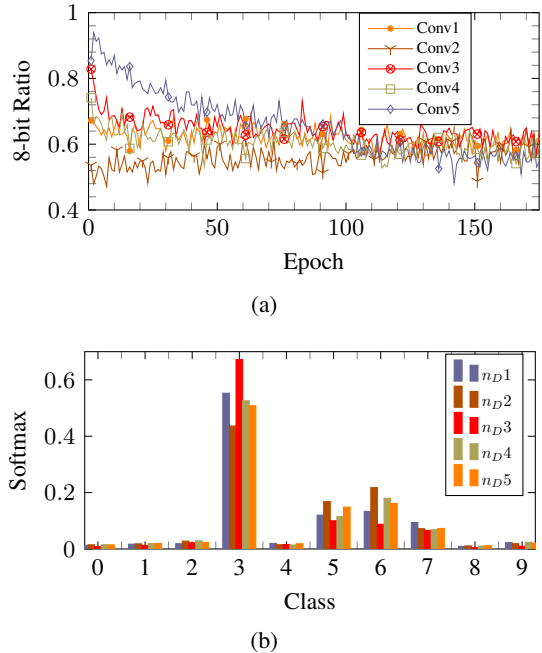

\subsection{Stochastic Ensemble Learning}
\label{ch4:subsec:ensemble}

Intuitively, using a good teacher when training QDNN with KD will improve the performance~\cite{tang2020understanding}. In the proposed method, the activation quantization precision, $\mathbf{n}_{\textrm{SPP}}$, for each layer was determined stochastically. In this case, the total number of combinations for $\mathbf{n}_{\textrm{SPP}}$ is $2^L$, and among them there will be the solution, $\mathbf{n}_{\textrm{SPP}}^*$, which shows the best performance. However, finding this solution is not practical for DNNs because $2^L$ inferences are needed for the exhaustive search. To investigate the performance of the best solution, we design a shallow CNN that consists of five convolutional layers and train the model by selecting the $\mathbf{n}_{\textrm{SPP}}$ according to the true label, $y$, for each step as follows:
\begin{align}
    \mathbf{n}_{\textrm{SPP}}^{\ast} = \argmin_{\mathbf{n}}\mathcal{L}_{CE} (y, \mathbf{z}_{\textrm{SPP}} | \mathbf{n}). \label{ch4:eq:greedy}
\end{align}
Since the experiment is performed on a five-layer CNN, $L=5$, we employ the greedy strategy that finds the $\mathbf{n}_{\textrm{SPP}}^*$ by inferencing the model $2^5$ times with different combinations of the quantization precision. Note that $\mathbf{n}_{\textrm{SPP}}^*$ can change for each training step. The target model is trained using the soft label obtained with $\mathbf{n}_{\textrm{SPP}}^*$. \figurename ~\ref{ch4:fig:ratio_change_curve} (a) shows how the ratio of 8-bit selection changes during training for each layer. The model is pretrained to the 2-bit weights and activations and the target precision is also 2 bits. The floating-point and 2-bit models show accuracies of 89.9\% and 87.8\%, respectively. 

The solution of Eq.~\eqref{ch4:eq:greedy} is not always 8-bit even at the beginning of the training. Note that the results in~\tablename~\ref{ch4:table:lowtohighex} show that using higher precision for the activation can achieve higher average accuracy. For each iteration, however, choosing 8-bit activation may not show the lowest loss. More importantly, the ratio of 8-bit selection decreases to 0.6 as the training progresses. This indicates that the best-performing solution, $\mathbf{n}_{\textrm{SPP}}^{*}$, selects 2- and 8-bit almost uniformly. As a result, the test accuracy of the 2-bit model with greedy training is 88.4\%, which is a better performance of always choosing 8-bit, 88.1\%.

Another advantage of the SPEQ is that it has the effect of ensemble learning. By the stochastic selection of bit precision, soft labels with different distributions can be created for one training sample. In this case, the diversity of soft labels should be large enough to obtain the effect of the ensemble well~\cite{chen2019online}. \figurename~\ref{ch4:fig:ratio_change_curve} (b) shows the computed soft labels by quantizing the activation outputs of the ResNet20 with different bit precisions for a single training sample in the CIFAR10 dataset. Although we extract soft labels from the same parameters, the distribution of the soft labels varies according to the activation precision.


Based on our analysis, we apply the SPEQ training method with uniform $n_{A}$-bit and 8-bit selection probabilities to increase diversity. The sensitivity of bit-precision candidate or the quantization probability, $u$, is also examined.


\begin{table*}[t]
\setlength\tabcolsep{8pt}
\caption{Comparison of the test accuracy of 2-bit ResNet20 on CIFAR10 according to the loss function for KD. The cosine similarity loss (CS-Loss) suits better than the KL-divergence loss (KL-Loss) for the proposed SPEQ method. Average test accuracy of 5 repeated experiments is reported with the standard deviation.}
\label{ch4:table:cosloss}
\begin{center}
\begin{tabular}{l|cc}
\toprule
Method (2-bit baseline accuracy: 90.73\%)                                                          & KL-Loss   & CS-Loss \\ \midrule
KD w/ full precision ResNet20 as teacher        & 91.24$\pm$0.06     & 91.22$\pm$0.10    \\
SPEQ ($u=0$, always choose 8-bit for soft labels) & 91.22$\pm$0.16 & 91.18$\pm$0.07    \\
SPEQ ($u=0.5$, 2-bit or 8-bit for soft labels)    & 90.83$\pm$0.07     & 91.44$\pm$0.04    \\
\bottomrule
\end{tabular}
\end{center}
\end{table*}

\begin{table*}[t]
\centering
\setlength\tabcolsep{2.6pt}
\caption{2-bit ResNet20 test accuracy according to the quantization probability for the stochastic precision path, $u$. Average test accuracy of 5 repeated experiments is reported. The result of `Mix' is obtained by selecting all precisions from 2 to 8 bits uniformly.}
\label{ch4:tab:u_change}
\begin{tabular}{c|cccccccccccc}
\toprule
$u$           & 0.0   & 0.1   & 0.2   & 0.3   & \textbf{0.4}   & \textbf{0.5}   & \textbf{0.6}   & 0.7   & 0.8   & 0.9   & 1.0   & Mix   \\ \midrule
Test Acc. & 91.22 & 91.23 & 91.22 & 91.29 & \textbf{91.39} & \textbf{91.44} & \textbf{91.43} & 91.21 & 91.24 & 90.96 & 90.74 & 91.23 \\ 
\bottomrule
\end{tabular}
\end{table*}




\begin{figure}[t]
\centering
   \subfigure[KL-Loss]{\includegraphics[width=0.49\linewidth,height=0.3\linewidth]{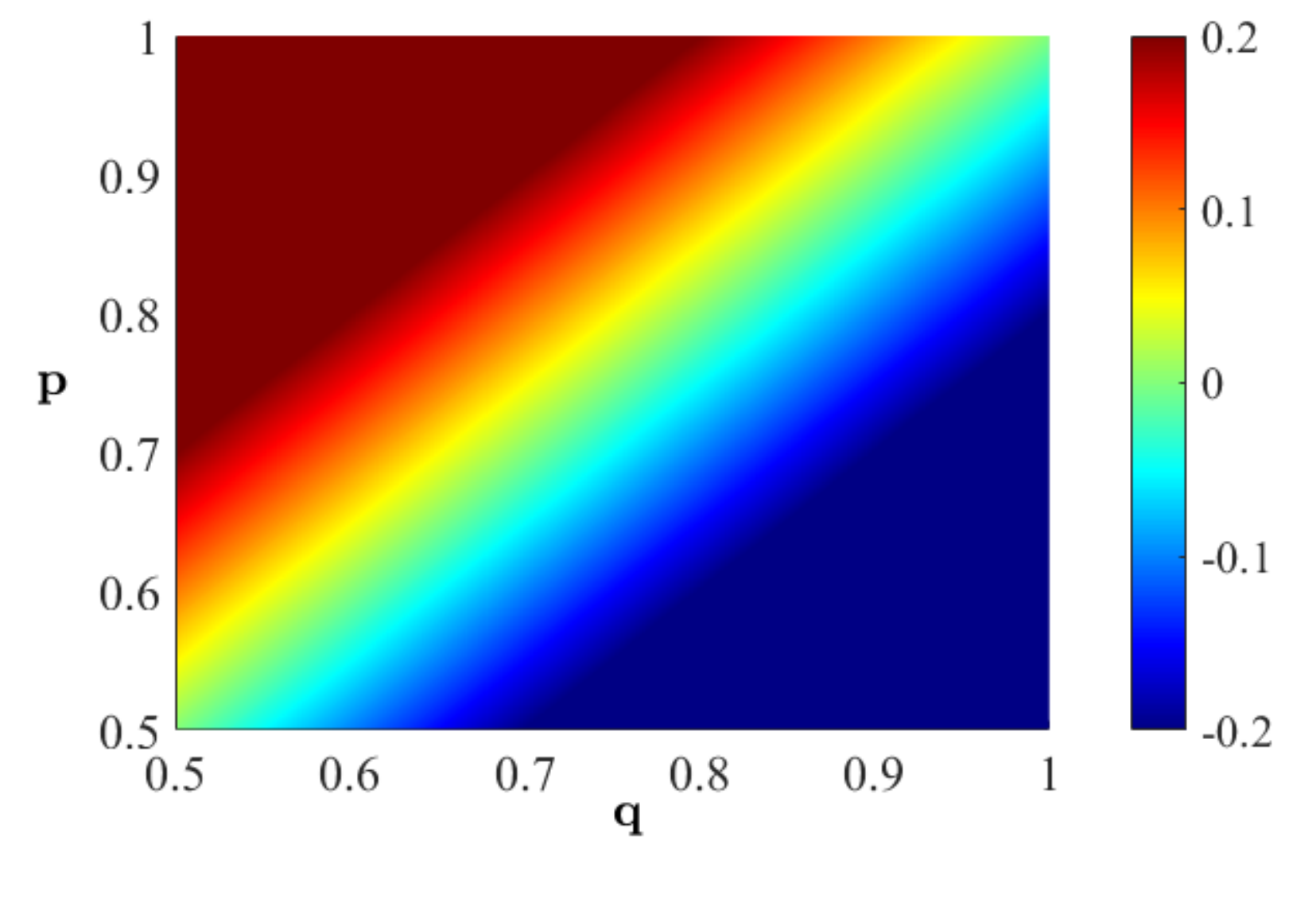}}
   \hfill
   \subfigure[CS-Loss]{\includegraphics[width=0.49\linewidth,height=0.3\linewidth]{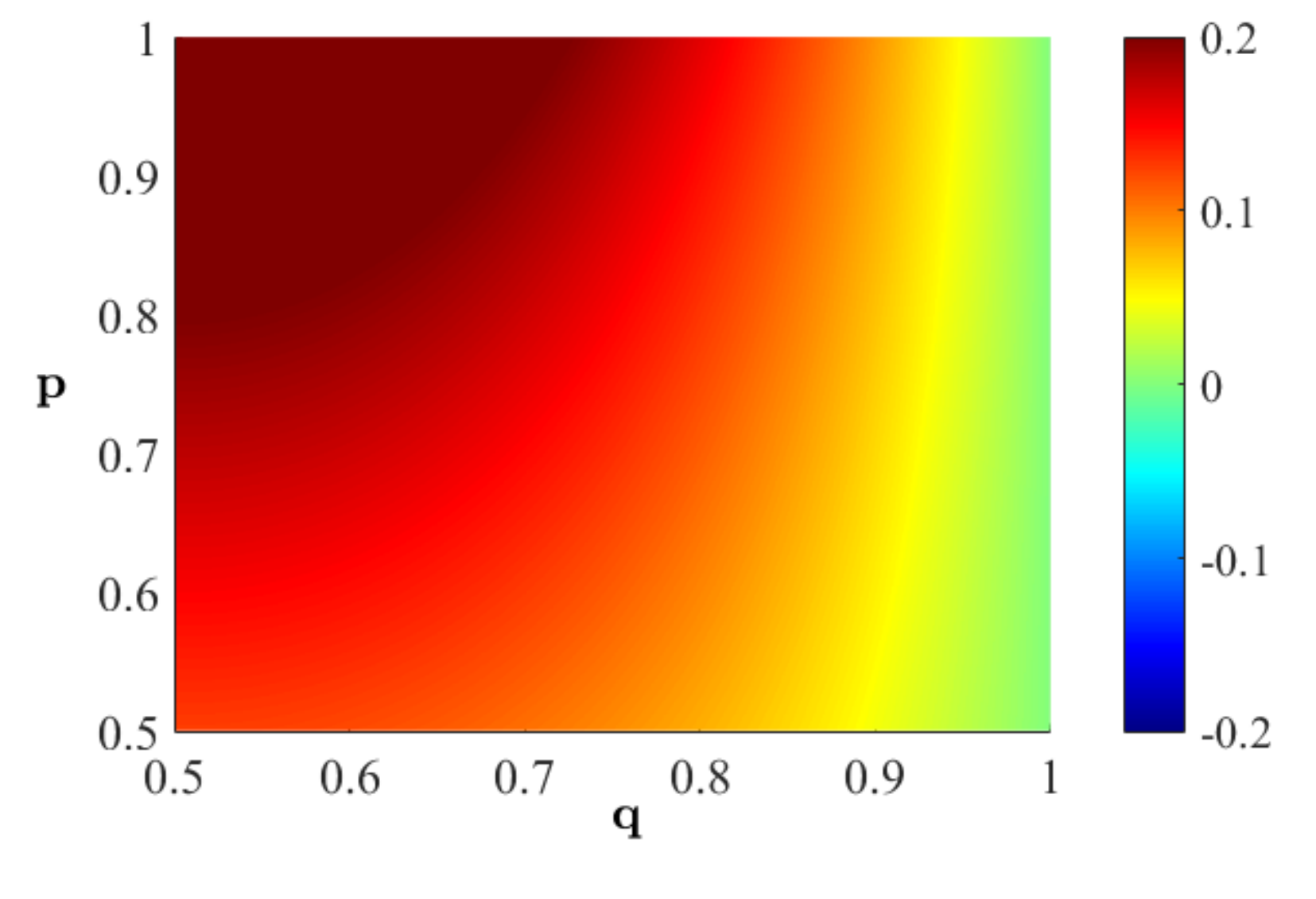}}
\caption{The gradients of the ground-truth logit according to the loss type. The x- and y-axes are the probabilities of the student and teacher for the logit, respectively.
} \label{ch4:fig:cosloss}
\end{figure}



\subsection{Cosine Similarity Learning}
\label{ch4:subsec:coslearing}

In many KD approaches, KL-divergence is commonly used as a loss function to reflect the guidance of the teacher. In the setting of SPEQ, however, we claim that the cosine similarity loss (CS-Loss) functions better than KL-divergence loss (KL-Loss). The main difference is that the teacher in SPEQ may not be more reliable than the student. Note that activations are randomly quantized in SPEQ, thus the output prediction of the teacher might be significantly affected by the quantization noise. In this setting, it is important to reflect the guidance of the teacher selectively, as there is no guarantee that the teacher's prediction is more accurate than the student's. In this section, we explain that CS-Loss has such capability whereas KL-Loss does not. 

To understand the situation more concretely, we compare the back-propagation errors (i.e., gradients w.r.t. each logit) for the two loss functions. Note that the student model is guided to increase (or decrease) the logit if the corresponding gradient is negative (or positive).
When the predictions of the teacher and the student are $\mathbf{p}$ and $\mathbf{q}$, respectively, the KL-Loss and its back-propagation error for the $i^{\textnormal{th}}$ logit, $z_{i}$, is represented as follows~\cite{hinton2015distilling}:
\begin{align}
    \mathcal{L}_{KL} (\mathbf{p},\mathbf{q}) &= -\sum_{i=0}^{C-1} p_{i} \log \frac{q_{i}}{p_{i}}, \\
\frac{\partial \mathcal{L}_{KL}}{\partial z_{i}} &= q_{i} - p_{i}. \label{ch4:eq:dkl}
\end{align}
Eq.~\eqref{ch4:eq:dkl} indicates that the $\mathcal{L}_{KL}$ produces back-propagation errors in the direction of decreasing the difference between the $p_{i}$ and $q_{i}$. That is, KL-Loss guides the student to always follow the teacher. Such guidance is regarded as "re-weighting" ~\cite{tang2020understanding}, but it is helpful under a condition that the teacher's prediction is more confident than the student's. Since the teacher in SPEQ is not as reliable as the large teacher models in typical KD methods, the gradients from KL-Loss can be misleading.


In comparison, CS-Loss between predictions of the teacher and the student after the normalization is given as follows:
\begin{align}
        \mathcal{L}_{CS} (\mathbf{p},\mathbf{q}) &= 1 - \mathbf{p} \cdot \mathbf{q}, \\
    \frac{\partial \mathcal{L}_{CS}}{\partial z_{i}} &= -\sum_{j=0}^{C-1} p_{j} (q_{j} \delta_{ij} - q_{j} q_{i}). \label{ch4:eq:dcos}
\end{align}
The gradients of $\mathcal{L}_{CS}$ are more cognizant of the confidence of both the teacher and the student. Assume that the $i^{\textnormal{th}}$ label is the ground-truth and the teacher’s prediction is also confident about it, i.e. $p_{i} >> p_{j}$ ($i\neq j$), Eq.~\eqref{ch4:eq:dcos} is approximated as follows:
\begin{align}
    \frac{\partial \mathcal{L}_{CS}}{\partial z_{i}} \approx -p_{i}q_{i}(1-q_{i}). \label{ch4:eq:dcos_dom}
\end{align}
Eq.~\eqref{ch4:eq:dcos_dom} indicates that the gradients is proportional to $q_{i}(1-q_{i})$. This is particularly helpful when the confidence of the student's prediction is not high; when $0<q_{i}<1$, the student is guided to increase the confidence for $q_{i}$. If the student's prediction has high confidence, the gradients become zero so that they will be highly penalized by the CE-Loss if the prediction is wrong. 


In addition, as the prediction of the teacher itself is ambiguous, the back-propagation error decreases. When the prediction of the teacher goes to a uniform distribution, Eq.~\eqref{ch4:eq:dcos} can be approximated as:
\begin{align}
    \frac{\partial \mathcal{L}_{CS}}{\partial z_{i}} \approx -p_{i}q_{i}(1-\sum_{j=0}^{C-1} q_{i}) = 0
\end{align}
Note that the gradients in this case are almost zero, implying that the CS-Loss will be neglected when the confidence of the teacher's prediction is small. Additional case-studies and the examples are provided in Appendix C.

The impact of the relationship between the teacher and the student predictions to the gradients is illustrated in~\figurename~\ref{ch4:fig:cosloss}. As can be seen, the gradient of KL-Loss flips its direction when the student's confidence is higher than the teacher. This is detrimental for SPEQ-based knowledge distillation as the prediction of the teacher is prone to noise. Whereas, the CS-Loss allows selective adoption of the teacher's information; the gradients guide to follow the teacher more if it has high confidence. If not, the guidance is neglected.  

To distinguish the effects of two loss terms on general and our KD, we trained the 2-bit ResNet20 using various teachers as shown in~\tablename~\ref{ch4:table:cosloss}. The baseline model is trained without applying KD. When the output probability of the teacher is computed using a better teacher model or a deterministic precision on the shared model, the two loss terms show almost the same result. However, when the teacher outputs are generated while changing the bit precision, only the cosine loss shows the performance improvement compared to the baseline model. 

\begin{table*}[t]
\setlength\tabcolsep{8pt}
\caption{Test accuracy (\%) of quantized CNNs on CIFAR10 and CIFAR100 datasets. `F' denotes the floating-point precision.}
\label{ch4:tab:cifar10_100}
\begin{center}
\begin{tabular}{lc|cc|cc}
\toprule
\multirow{2}{*}{Methods} & Precision     & \multicolumn{2}{c|}{CIFAR10} & \multicolumn{2}{c}{CIFAR100} \\
                         & (W / A) & VGG16       & ResNet20      & ResNet32    & MobileNetV2    \\ \midrule
Baseline                 & F / F         & 93.6        & 92.1          & 70.3        & 76.8           \\ \midrule
Retrain             & 2 / 2         & 92.5        & 90.7          & 66.9        & 73.0           \\
PACT-SWAB-8brc~\cite{choi2019accurate}        & 2 / 2         & -           & 90.7          & -           & -              \\
QKD~\cite{kim2019qkd}                   & 2 / 2         & -           & 90.5          & 66.4        & -              \\
SPEQ                     & 2 / 2         & \textbf{93.1}        & \textbf{91.4}          & \textbf{69.1}        & \textbf{74.4}           \\ \midrule
Retrain             & F / 2         & 92.9        & 91.8          & 67.9      & 74.5         \\
SPEQ                     & F / 2         & \textbf{93.5}        & \textbf{92.1}          & \textbf{69.7}      & \textbf{75.2}        \\ \bottomrule
\end{tabular}
\end{center}
\end{table*}

\begin{table*}[t]
\setlength\tabcolsep{5pt}
\caption{Top-1 validation accuracy (\%) on the ImageNet dataset. Values in the parentheses are the accuracy of pretrained floating-point models.}
\label{ch4:tab:imagenet}
\begin{center}
\begin{tabular}{lccc|lcc}
\toprule
\multicolumn{4}{c|}{2-bit weights / 2-bit activations}                                                                                                                                               & \multicolumn{3}{c}{Float weights / 2-bit activations}                                                                                                                                             \\ \midrule
\multicolumn{1}{l}{Method}        & \begin{tabular}[c]{@{}c@{}}AlexNet \\ (60.8)\end{tabular} & \begin{tabular}[c]{@{}c@{}}ResNet18 \\ (70.3)\end{tabular} & \begin{tabular}[c]{@{}c@{}}ResNet34 \\ (73.6)\end{tabular} & Method       & \begin{tabular}[c]{@{}c@{}}AlexNet \\ (60.8)\end{tabular} & \begin{tabular}[c]{@{}c@{}}ResNet18 \\ (70.3)\end{tabular}   \\ \midrule
DoReFa$^\dagger$~\cite{zhou2016dorefa}   &   46.4 &  62.6    & - & BalancedQ~\cite{zhou2017balanced}  & 56.5  & 61.1   \\
QIL~\cite{jung2019learning}        & 58.1    & 65.7  & 70.6  & QN~\cite{yang2019quantization}        & -  & 65.7      \\
PACT\_SWAB~\cite{choi2019accurate} & 57.2    & 67.0   & -     & DoReFa$^\dagger$~\cite{zhou2016dorefa}    & 54.1  & 66.9       \\
Retrain    & 56.9    & 66.6  & 70.5  & PACT~\cite{choi2018pact}      & 54.9        & 67.5                 \\
SPEQ          & \textbf{59.3}  & \textbf{67.4} & \textbf{71.5}    & SPEQ         & \textbf{60.8} & \textbf{68.4} \\ \bottomrule
\end{tabular}
\end{center}
\scriptsize{Results with the symbol $^\dagger$ are from \cite{choi2018pact}.}

\end{table*}

\section{Experimental Results}
\label{ch4:sec:exp}


The training procedures in our experiments consist of three steps: train the floating-point DNN (pretrain), train the QDNN to the target precision initialized from the floating-point parameters (retrain~\cite{hwang2014fixed}), and train the QDNN using the SPEQ method initialized with the retrained parameters. 
The details of the experimental settings for each task explained in Appendix D.

\subsection{Results on the CIFAR10 and CIFAR100 Datasets}

We first studied how the stochastic quantization probability, $u$, affects the performance of the SPEQ method. To this end, we trained 2-bit quantized ResNet20 models with various values of $u$ from 0.0 to 1.0 on the CIFAR10 dataset. The results are reported in \tablename~\ref{ch4:tab:u_change}. It should be noted that 0.0 and 1.0 of $u$ indicate that the stochastic precision path selects only 8- and 2-bit precisions, respectively. The best test accuracy is observed when $u$ is between 0.4 and 0.6. This result indicates that the SPEQ shows the best performance when the stochastic precision path is selected to some degree evenly rather than being biased to either precision. For comparison, we employed SPEQ by uniformly selecting all precisions from 2 to 8 bits. The result is shown as `Mix’ in \tablename~\ref{ch4:tab:u_change}. The result is 0.21\% lower than that of the training using only two precisions, 2 and 8 bits. Note that Quantization errors are the largest in 2-bit precision and the lowest in 8-bit precision. Rather than intermediate precision, selecting precision either the lowest or highest precision increases the diversity of teacher outputs, which has a great influence on performance~\cite{chen2019online}. The uniform selection between the lowest and highest precision results in a better ensemble effect by increasing the diversity of teacher outputs. For all the rest of the experiments, we set the quantization probability, $u$, to 0.5.

We evaluated the proposed SPEQ scheme using the CIFAR10 and CIFAR100 datasets. The performance of the SPEQ and existing methods are shown in~\tablename~\ref{ch4:tab:cifar10_100}. The test accuracy of 2-bit ResNet20 before applying SPEQ, denoted as `Retrain', is 90.7\% on the CIFAR10 dataset. The SPEQ significantly improves the performance of 2-bit ResNet20 and achieves 91.4\% test accuracy. This result is better than the QKD~\cite{kim2019qkd}, which employs a large teacher. Furthermore, when only activations are quantized to 2 bits, SPEQ shows almost the same performance as the full-precision models for CIFAR10. The SPEQ shows consistent improvements on various CNNs.

\subsection{Results on the ImageNet Dataset}

The performance of the SPEQ on the ImageNet dataset is shown in~\tablename~\ref{ch4:tab:imagenet}. The retraining scheme shows 56.9\%, 66.6\%, and 70.5\% top-1 accuracy for the 2-bit AlexNet, ResNet18, and ResNet34, respectively. By SPEQ training, the top-1 accuracy increases approximately 1\% for ResNet18 and ResNet34. The SPEQ training on 2-bit AlexNet improves the top-1 accuracy noticeably, showing 59.3\% top-1 accuracy. This result is only a 1.5\% accuracy drop compared to the full-precision AlexNet. The results for 2-bit activation-quantized CNNs indicate that the proposed SPEQ method is very effective for reducing the activation quantization noise.

We also optimized the 3-bit EfficientNet-b0 with the ImageNet dataset. We obtained the pretrained full-precision model from the Tensorflow official Github~\footnote{\url{https://github.com/tensorflow/tpu/tree/master/models/official/efficientnet}} and retrained the model using the same hyper-parameters and data augmentation methods as those for ResNet. The performance of 3-bit EfficientNEt-b0 is compared in~\tablename~\ref{ch4:tab:imagenet_efficientnet}. Note that the AP~\cite{mishra2018apprentice} and QKD~\cite{kim2019qkd} employed the full-precision EfficientNet-b1 as a teacher for KD. SPEQ outperformed the existing KD methods on this recently developed CNN without a large teacher model. 

\begin{table}[t]
\caption{ImageNet valiation top-1 accuracy on 3-bit weight and activation quantized EfficientNet-b0. The top-1 accuracy of the full-precision model is 76.7\%.}
\label{ch4:tab:imagenet_efficientnet}
\begin{center}
\begin{tabular}[t]{c|cccc}
    \toprule
 Methods & Retrain     & SPEQ & AP$^\dagger$ & QKD$^\dagger$  \\ \midrule
Top-1 Acc (\%) & 68.4 & \textbf{69.5} & 68.4 & 69.2 \\ \bottomrule
\end{tabular}
\end{center}
\scriptsize{Results with the symbol $^\dagger$ are from \cite{kim2019qkd}.}
\end{table}

\begin{table}[t]
\setlength\tabcolsep{2.1pt}
\caption{2-bit ImageNet quantization Top-1 validation accuracy (\%) compared with other KD applied QDNNs.}
\label{ch4:tab:imagenet_kd}
\begin{center}
\begin{tabular}{llc|lc}
\toprule
Method   & \multicolumn{2}{c|}{ResNet18} & \multicolumn{2}{c}{ResNet34} \\
               & Teacher         & Acc(\%)    & Teacher          & Acc(\%)     \\ \midrule
Retrain & w/o KD     & 66.6     & w/o KD   & 70.5     \\ \midrule
AP$^\dagger$ & ResNet34 & 66.8     & ResNet50     & 71.1     \\
QKD$^\dagger$         & ResNet34 & 67.4        & ResNet50     & 71.6    \\
SPEQ         & -      & 67.4         & -     & 71.5          \\ 
SPEQ+AP           & ResNet34       & \textbf{67.8}         & ResNet50    & \textbf{72.1}          \\ \bottomrule
\end{tabular}
\end{center}
\scriptsize{``Acc'' with the symbol $^\dagger$ are from \cite{kim2019qkd}.}
\end{table}


\begin{table}[t]
\setlength\tabcolsep{10pt}
\caption{The performance (EM and F1 scores) of low-precision BERT on the SQuAD1.1 dev dataset. FixedBERT results are from~\cite{boo2020fixed}}
\label{ch4:tab:bert}
\begin{center}
\begin{tabular}{l|cc|cc}
\toprule
\multirow{2}{*}{SQuAD1.1}                            & \multicolumn{2}{c|}{W3/A3} & \multicolumn{2}{c}{W4/A4} \\
& EM     & F1     & EM     & F1     \\ \midrule
FixedBERT        &  71.5 & 81.4  & 74.2     & 83.1     \\
SPEQ                & \textbf{76.4} & \textbf{85.1} & \textbf{78.0} & \textbf{86.6} \\ \bottomrule
\end{tabular}
\end{center}
\end{table}

\begin{table}[t]
\setlength\tabcolsep{4pt}
\caption{Validation accuracy (\%) on the Flowers-102 dataset according to the training method and the feature extractor. RN is an abbreviation for ResNet.}
\label{ch4:tab:flower}
\begin{center}
\begin{tabular}{l|ccc}
\toprule
\# training & Float RN18 & 2-bit RN18 & 2-bit RN18 \\
samples        & (CE Loss)        & (CE Loss)        & (SPEQ) \\ \midrule
510 (5 / label)  & 69.18            & 70.67          & \textbf{71.21}                \\
1020 (10 / label)  & 78.04           & 77.99           & \textbf{78.36}                \\
2040 (20 / label) & 84.57           & 83.91           & \textbf{85.02}                \\
 \bottomrule
\end{tabular}
\end{center}
\end{table}

The key difference in the training procedure between SPEQ and previous works is that SPEQ shares the same model for the teacher and the student. This simple choice leads to the significant savings in the training computation. Although our approach is based on the self-distillation, the larger teacher can also be employed to further improve the performance of the target model. The SPEQ training with the KD method is also compared with other KD training methods for QDNNs in~\tablename~\ref{ch4:tab:imagenet_kd}. We applied the KD by combining the Apprentice~\cite{mishra2018apprentice} (AP) and the SPEQ scheme. The combined training improves the top-1 accuracy of ResNet18 and ResNet34 by 0.4\% and 0.6\%, respectively.



\subsection{Results on Transfer Learning}

Because the proposed method requires little computational overhead, it can be applied to transfer learning from a very large model such as BERT~\cite{devlin2018bert}. We optimized the low-precision BERT using SPEQ training. The pretrained BERT-Base model is obtained from the Google research~\footnote{\url{https://github.com/google-research/bert}} and fine-tuned using the Stanford Question Answering Dataset (SQuAD1.1)~\cite{rajpurkar2016squad}. The performance improvements of the SPEQ on the quantized BERT are shown in~\tablename~\ref{ch4:tab:bert}. The fine-tuned floating-point BERT shows 81.1 F1 and 88.6 EM scores. When the activation is quantized to $n_{A}$ bits, the stochastic precision for computing soft labels is chosen between $n_{A}$ and 8 bits. 

We expand the experiment for transfer learning using Oxford Flowers-102~\cite{nilsback2008automated}. We employ the ResNet18 as a feature extractor, which is frozen when fine-tuning. The SPEQ is evaluated by changing the number of training samples and the results are shown in~\tablename~\ref{ch4:tab:flower}. SPEQ improves the performance significantly compared to the 2-bit model retraining using the cross-entropy loss. Moreover, the SPEQ-trained 2-bit ResNet18 achieves better results than the floating-point model. In practice, the conventional KD method is hard to be applied due to the need of auxiliary models. Therefore, the SPEQ method is an invaluable solution to apply KD on transfer learning.

\section{Concluding Remarks}
In this work, we proposed a novel KD method for quantized DNN training. The proposed method, SPEQ, does not require a cumbersome teacher model; it assigns the same parameters for the teacher and student networks. The teacher model is formed by assigning the stochastic precision to the activation of each layer, by which it can produce the soft labels of stochastically ensembled models. The cosine similarity loss is used for KD training to render reliable operation even when the confidence of the teacher is lower than that of the student. The SPEQ outperforms the existing quantized training methods in various tasks. Furthermore, the SPEQ can be easily used for low-precision training even when no larger teacher model is available.

\bibliography{refs}

\clearpage

\section*{Supplementary Materials}

\subsection*{A. Implementation details for high-precision inference}

We describe the details of increasing the precision for inference in~\tablename 1 of the main paper. We conducted a simple experiment on the ResNet20 model using the CIFAR100 dataset. ResNet20 model was trained with either weight-only or activation-only quantization (into 2-bit). The quantization method we employed is described in Section B. The two quantized models (W2AF or WFA2) are then employed for inference, where equal or higher bit precision (2 to 8-bit) is used. When increasing the precision of activation or weights, the same clipping level is employed. Thus, the number of discretization points increases while the representation range is consistent. To increase the discretization points of 2-bit weights, we quantize the full-precision weights that are used to accumulate the gradients~\cite{courbariaux2015binaryconnect,hwang2014fixed}. We assume not to abandon any quantization points.

\subsection*{B. Quantization method}

In this section, we introduce the quantization method used in this study. We employ a layer-wise uniform symmetric quantizer for efficient implementations. This quantizer consists of clipping and quantization to limit the range and precision of variables. The weights or the input signals in the same layer share a trainable scalar clipping value as suggested in~\cite{choi2018pact,jung2019learning}.

We employed PACT~\cite{choi2018pact} for the $n$-bit quantization of rectified linear unit (ReLU) as follows:
\begin{align}
\hat{x} &= 0.5(|x| - |x-\alpha_x| + \alpha_x) \\
Q(x) &= \lceil \hat{x} \cdot (\frac{2^{n}-1}{\alpha_x}) \rfloor \frac{\alpha_x}{2^{n}-1},
\end{align}
where $\lceil \cdot \rfloor$ is the rounding operation. For example, 2-bit quantized activation output is one of $\{0, \alpha_x / 3, 2\alpha_x /3 , \alpha_x\}$. Gradients for  $\lceil \cdot \rfloor$ are calculated using the straight-through estimator (STE)~\cite{bengio2013estimating} and $\alpha_x$ is trained according to the following back-propagated gradients:
\begin{align}
\frac{\partial L}{\partial \alpha_x} = \frac{\partial L}{\partial Q(x)} \frac{\partial Q(x)}{\partial \alpha_x} = 
\begin{cases}
\frac{\partial L}{\partial Q(x)} & x > \alpha_x \\
0 & \textrm{else},
\end{cases}
\end{align}
When all activation outputs are less than $\alpha_x$, the gradients for $\alpha_x$ becomes 0. Therefore, L2 regularization is applied to decrease the $\alpha_x$ when all activation output values are smaller than the $\alpha_x$.

For $n$-bit weight quantization, we slightly modify the PACT algorithm as follows:
\begin{align}
\hat{w} &= 0.5(|w + \alpha_w| - |w-\alpha_w|) \\
\hat{w}' &= \frac{\hat{w}}{2 \alpha_w} + 0.5 \\
Q(w)' &= \lceil \hat{w}' \cdot (2^{n}-1) \rfloor / (2^{n}-1) \\
Q(w) &= 2 \alpha_w (Q(w)' - 0.5),
\end{align}
where the $w$ is the weights, $\alpha_{w}$ is the clipping value for the weights in a layer. For example, the 2-bit weights employ four levels, which are $\{-\alpha_w$, $- \alpha_w /3$,  $\alpha_w /3$, $\alpha_w \}$. Similar to the PACT activation quantization, the gradients for $\alpha_w$ becomes:
\begin{align}
\frac{\partial L}{\partial \alpha_w} = 
\begin{cases}
\frac{\partial L}{\partial Q(w)} & x > \alpha_x \\
- \frac{\partial L}{\partial Q(w)} & x < -\alpha_x \\
0 & \textrm{else.}
\end{cases}
\end{align}

\begin{figure*}[h]
\centering
  \subfigure[]{
\begin{tikzpicture}
    \begin{axis}[ylabel shift = 2.5pt,
	ybar=0pt,
    width=0.4\linewidth,
	height = 0.2\linewidth,
	bar width=2.5pt,
    xmin=0,
	ymin=0,
	xmax=9,
	ymax=0.65,
    label style={font=\footnotesize},
    legend style={nodes={scale=0.62, transform shape},at={(0.95,0.95)},anchor=north east},
    xlabel={Class},
    tick label style={font=\footnotesize}, 
	minor x tick num=1, 
	enlarge x limits={abs=2*\pgfplotbarwidth},
	xtick pos=both, 
	xtick align=inside, 
	major tick style={line width=0.010cm, black},
	major tick length=0.10cm,
	nodes near coords align={vertical},
    symbolic x coords={0,1,2,3,4,5,6,7,8,9} ,
        ylabel=Softmax]
	\legend{$\mathbf{p}$ (teacher), $\mathbf{q}$ (student)};
    \addplot[color=blue!60!yellow,fill=blue!60!yellow] file{chapter3_sources/b1.txt}; 
	\addplot[color=red!70!green,fill=red!70!green] file{chapter3_sources/b2.txt}; 
    \end{axis}
  \end{tikzpicture}}
  \hskip 0.1in
  \subfigure[]{
\begin{tikzpicture}
    \begin{axis}[
    ybar=0pt,
    width=0.4\linewidth,
	height = 0.2\linewidth,
	bar width=2.5pt,
    xmin=0,
	ymin=0,
	xmax=9,
	ymax=0.65,
    label style={font=\footnotesize},
    legend style={nodes={scale=0.62, transform shape},at={(0.5,0.4)},anchor=north west},
    xlabel={Class},
    tick label style={font=\footnotesize}, 
	minor x tick num=1, 
	enlarge x limits={abs=2*\pgfplotbarwidth},
	xtick pos=both, 
	xtick align=inside, 
	major tick style={line width=0.010cm, black},
	major tick length=0.10cm,
	nodes near coords align={vertical},
    symbolic x coords={0,1,2,3,4,5,6,7,8,9}]
	
    \addplot[color=blue!60!yellow,fill=blue!60!yellow] file{chapter3_sources/d1.txt}; 
	\addplot[color=red!70!green,fill=red!70!green] file{chapter3_sources/d2.txt}; 
    \end{axis}
  \end{tikzpicture}}
  \subfigure[]{
\begin{tikzpicture}
    \begin{axis}[ylabel shift = -6.5 pt,
	width=0.4\linewidth,
	height = 0.2\linewidth,
	xmin=0,
	ymin=-0.51,
	xmax=9,
	ymax=0.5,
	xlabel={Class},
	label style={font=\footnotesize},
	legend style={nodes={scale=0.62, transform shape},at={(0.95,0.05)},anchor=south east},
	tick label style={font=\footnotesize}, 
	minor x tick num=1, 
	xtick pos=both, 
	xtick align=inside, 
	major tick style={line width=0.010cm, black},
	 major tick length=0.10cm,
        ylabel=Gradients]
	\legend{Cosine, KL};
    \addplot[color=blue!60!yellow, mark=*, mark size=1.0pt, solid, mark repeat=1,mark options=solid] file{chapter3_sources/a1.txt}; 
	\addplot[color=red!70!green, mark=Mercedes star flipped, mark size=2.5pt, solid, mark repeat=1,mark options=solid] file{chapter3_sources/a2.txt}; 
    \end{axis}
  \end{tikzpicture}}
  \subfigure[]{
\begin{tikzpicture}
    \begin{axis}[
	width=0.4\linewidth,
	height = 0.2\linewidth,
	xmin=0,
	ymin=-0.51,
	xmax=9,
	ymax=0.5,
	xlabel={Class},
	label style={font=\footnotesize},
	legend style={nodes={scale=0.62, transform shape},at={(0.85,0.99)},anchor=north west},
	tick label style={font=\footnotesize}, 
 	minor x tick num=1, 
	xtick pos=both, 
	xtick align=inside, 
	major tick style={line width=0.010cm, black},
	 major tick length=0.10cm]
	
    \addplot[color=blue!60!yellow, mark=*, mark size=1.0pt, solid, mark repeat=1,mark options=solid] file{chapter3_sources/c1.txt}; 
	\addplot[color=red!70!green, mark=Mercedes star flipped, mark size=2.5pt, solid, mark repeat=1,mark options=solid] file{chapter3_sources/c2.txt}; 
    \end{axis}
  \end{tikzpicture}}
\caption{Example of (a) softmax outputs and (c) gradients for the logits of the student when the teacher is more confident than the student. (b,d) Another example of softmax outputs and gradients when the teacher is less confident than the student.}
\label{ch4:fig:cosloss1}
\end{figure*}
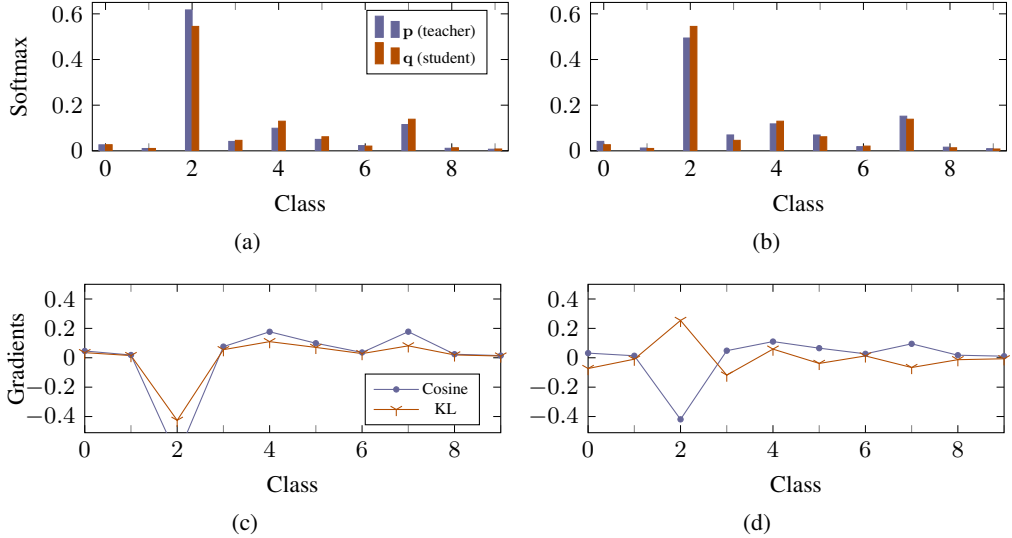

\subsection*{C. Gradient analysis of the cosine similarity loss}


The proposed SPEQ method computes a teacher output with randomly selected activation precision. In this case, a bad teacher outputs, i.e., less confident to the ground-truth, can be generated. \figurename~\ref{ch4:fig:cosloss1} (a) and (b) show examples of softmax outputs computed with different activation precision for one training sample. Those examples are the inference results on 2-bit ResNet20 for the sample where the ground-truth label is 2. The probability of the student, $q$, is calculated with 2-bit activation through the target precision path (TPP). Teacher probability, $p$, is the softmax output computed through the stochastic precision path (SPP). Depending on the activation precision, the confidence of the teacher outputs for the ground-truth can be higher or lower than the student output. When more confident teacher output is selected, the gradients for the student's logits are shown in~\figurename~\ref{ch4:fig:cosloss1} (c). Conversely, gradients with a less confident teacher output are shown in~\figurename~\ref{ch4:fig:cosloss1} (d).

The cosine similarity loss (CS-Loss) and KL-divergence loss (KL-Loss) produce similar gradients when the teacher's confidence in the ground-truth is higher than the student's. However, when the confidence of the teacher is lower than that of the student, KL-Loss creates a positive gradient for the logit corresponding to the ground-truth. Note that this gradient lowers the student's logit for the ground-truth. Therefore, the direction of gradients changes depending on the selected teacher. This hinders training the student model in a consistent direction. In results, the effects of KD diminish with the proposed SPEQ method when using the KL-Loss. Experimentally, applying KL-Loss to SPEQ (90.83\%) showed similar performance compared to the training without KD (90.73\%). The results in the parentheses are shown in Table 2 of the main contents.

On the other hand, the direction of the gradient for the ground-truth logit does not change when CS-Loss is employed. This is because the teacher's probability acts as a scaling factor. When the confidence of the teacher is small, the gradient for the ground-truth logit also becomes small as shown in~\figurename~\ref{ch4:fig:cosloss1} (d). In results, employing CS-Loss instead of the KL-Loss showed much better performance for the proposed SPEQ method.

\subsection*{D. Experimental details}

\textbf{CIFAR10/CIFAR100}: We assess the proposed method on VGG16,  ResNet models, and MobileNetV2. VGG16 and ResNet20 are trained using the CIFAR10 dataset. For CIFAR100 dataset, ResNet32 and MobileNetV2 are employed. Training images are augmented by horizontally flipping and cropping~\cite{lee2015deeply}. All models for CIFAR10 and CIFAR100 datasets are trained using the same optimizer and hyper-parameters. The SGD optimizer is used with the momentum factor of 0.9 and the batch size is 128. We first trained full-precision models with the initial learning rate of 0.1. The leaning rate decays by the factor of 0.1 at 100 and 150 epochs. The total number of training epochs is 175. L2-loss is applied to the scale of 5e-4. The hyper-parameters for retraining and SPEQ methods are the same as follows. QDNNs are trained for 175 epochs with the initial learning rate of 0.01. The learning rate decreases by 0.1 times at 100 and 150 epochs. L2-loss is applied only for the activation clipping values with a scale of 5e-4. The hyper-parameters for SPEQ methods are the same as that of retraining. The temperature, $\mathcal{T}$, is set to 5.0 and 3.0 for the CIFAR10 and CIFAR100 datasets, respectively. The ReLU6 operation is used instead of ReLU for developing floating-point pretrained models. The activation clipping value $ \alpha_{x} $ is initialized to 6 for retraining. The weight clipping value $ \alpha_{w} $ is initialized to the value that minimizes L2-distance before and after the quantization of pretrained weights using Lloyd-algorithm~\cite{hwang2014fixed}. To reduce the variance in the training process, the 100 times lower learning rate is applied to $\alpha_{w}$~\cite{jung2019learning}.

\textbf{ImageNet}: For the ImageNet dataset, we evaluate our method on AlexNet, ResNet18 and ResNet34. The SGD optimizer is employed with the momentum of 0.9. The learning rate for the full-precision training is 0.4 with the batch size of 1024. The initial learning rate for low-precision training is 0.04. We train all the models for 90 epochs and the learning rate is decayed by a factor of 10 at 30, 50, 60, 70, and 80 epochs. The training images are augmented using random cropping and horizontal flipping. The input sizes for AlexNet and ResNet are $227\times227$ and $224\times224$, respectively. For the AlexNet, the batch-normalization is used instead of layer-normalization and we changed the position of max-pooling and activation layer to find the max value before quantization~\cite{rastegari2016xnor}. The ReLU6 is employed instead of ReLU when training floating-point models. The initial clipping values for quantization are obtained using the same method as the CIFAR10/CIFAR100 settings. The temperature, $\mathcal{T}$ is set to 1.0, 1.0, and 2.0 for AlexNet, ResNet18, and ResNet34, respectively.
 
\textbf{SQuAD1.1}:  We followed the quantization methods and hyperparameters for BERT proposed in~\cite{boo2020fixed}. All weights except the last output layer are quantized according to the weight quantization precision, $n_{W}$, and all hidden signals including the attention scores are quantized with the activation quantization precision, $n_{A}$. We fix the precision of attention scores to $n_{A}$ bits when computing soft labels because the stochastic quantization of attention scores rather decreases the performance in our experiments. Note that the attention-scores are always quantized to $n_{A}$ bits for the stochastic precision path (SPP). The temperature, $\mathcal{T}$, is set to 2.0 for all experiments.

\textbf{Flowers-102}: The Oxford Flowers-102 dataset consists of 8189 images the number of labels of 102. The number of images for each class varies from 40 to 258. With the consideration of the user-adaptation, we use a very small number of images as training samples, such as 5, 10, and 20 samples per label. The rest of the images are used as validation samples. All images are re-sized to 224$\times$224 and normalized channel-wisely. Note that data augmentation methods such as random cropping and flipping are not applied. We employed the ResNet18 model trained using the ImageNet dataset as a feature extractor and it is frozen when fine-tuning. The last output layer is newly added and trained for 50 epochs with a batch size of 16. The SGD optimizer is employed and the learning rate is 0.1.

\end{document}